\documentclass[conference,a4paper]{IEEEtran} 
\IEEEoverridecommandlockouts
\usepackage{cite}
\usepackage{amsmath,amssymb,amsfonts}
\usepackage{algorithmic}
\usepackage{graphicx}
\usepackage{textcomp}
\usepackage{xcolor}
\usepackage{subfig}
\usepackage[utf8]{inputenc}
\usepackage{array}

\def\BibTeX{{\rm B\kern-.05em{\sc i\kern-.025em b}\kern-.08em
    T\kern-.1667em\lower.7ex\hbox{E}\kern-.125emX}}

\begin{document}
\title{Beyond the Pipeline: Analyzing Key Factors in End-to-End Deep Learning for Historical Writer Identification}




\author{
    Hanif Rasyidi\IEEEauthorrefmark{1},
    Moshiur Farazi\IEEEauthorrefmark{3}\IEEEauthorrefmark{1} \\
    hanif.rasyidi@anu.edu.au\IEEEauthorrefmark{1}, 
    moshiur.farazi@udst.edu.qa\IEEEauthorrefmark{3} \\
    \IEEEauthorblockA{\IEEEauthorrefmark{1}College of Systems \& Society, 
    The Australian National University, Canberra, Australia}
    \IEEEauthorblockA{\IEEEauthorrefmark{3}Data Science and AI, 
    University of Doha for Science and Technology, Doha, Qatar}
}

\maketitle

\begin{abstract}

This paper investigates various factors that influence the performance of end-to-end deep learning approaches for historical writer identification (HWI), a task that remains challenging due to the diversity of handwriting styles, document degradation, and the limited number of labelled samples per writer. These conditions often make accurate recognition difficult, even for human experts. Traditional HWI methods typically rely on handcrafted image processing and clustering techniques, which tend to perform well on small and carefully curated datasets. In contrast, end-to-end pipelines aim to automate the process by learning features directly from document images. However, our experiments show that many of these models struggle to generalise in more realistic, document-level settings, especially under zero-shot scenarios where writers in the test set are not present in the training data. We explore different combinations of pre-processing methods, backbone architectures, and post-processing strategies, including text segmentation, patch sampling, and feature aggregation. The results suggest that most configurations perform poorly due to weak capture of low-level visual features, inconsistent patch representations, and high sensitivity to content noise. Still, we identify one end-to-end setup that achieves results comparable to the top-performing system, despite using a simpler design. These findings point to key challenges in building robust end-to-end systems and offer insight into design choices that improve performance in historical document writer identification.

\end{abstract}

\begin{IEEEkeywords}
historical document, writer identification, feature extraction, document retrieval
\end{IEEEkeywords}

\section{Introduction}

Writer identification (WI) is the process of analysing handwritten manuscripts to capture stylistic patterns unique to individual authors. Since handwriting is a form of behavioural biometrics, no two individuals produce identical writing styles, which can help with document classification \cite{rehman2019writer}. WI has been applied in diverse domains such as authorship attribution, signature forgery detection, and medical diagnostics, where handwriting analysis can help identify neurodegenerative diseases by tracking changes in a patient's writing over time. In the context of historical document analysis, WI plays a critical role in tracing document origin and enabling large-scale classification or clustering of unlabeled manuscripts.

Among these applications, historical writer identification (HWI) presents particularly difficult challenges due to the age of documents, the variability in handwriting, and the physical degradation of source materials. These factors make HWI one of the most complex and error-prone scenarios for both manual and automated analysis. Even for trained paleographers and forensic handwriting analysts, the task can be unreliable. External factors such as document damage often obscure important visual cues, while handwriting inconsistency may result from changes in ink, writing tools, environmental conditions, or script conventions \cite{seuret2020icfhr}. Experts typically rely on subtle visual characteristics such as stroke curvature, spacing, and letter formation, but their assessments can be subjective and inconsistent, especially when dealing with incomplete or degraded documents.

Efforts to automate HWI usually begin with pre-processing steps such as binarization and segmentation to isolate areas of interest (AOI). These are followed by handcrafted feature extraction techniques, including SIFT, HOG, oBIFs, or GLCM, and by measuring similarity through distance metrics or clustering algorithms. Although these methods have shown strong performance on well-curated datasets, they tend to be brittle and sensitive to noise. Their reliance on meticulous tuning and dataset-specific configurations often limits their scalability and generalisation, particularly in zero-shot scenarios involving unseen writers.

End-to-end deep learning offers a promising alternative by learning features directly from document images. With the rise of powerful convolutional and transformer-based backbones, several recent studies have applied such models to HWI tasks. Yet, their success is often limited to controlled scenarios such as word-level WI or datasets with similar text content between train and test splits. For example, some models rely on handcrafted AOI selection \cite{chammas2020writer} or require tightly constrained word-level annotations \cite{Kumar2024SiameseWriterID}, which do not generalise to document-level HWI. Other approaches enhance retrieval accuracy through re-ranking or pre-training on related domains \cite{Jordan2020, Cilia2019, Cilia2020}, but these methods are hard to reproduce and often entangle deep learning with domain-specific heuristics.

In this paper, we analyse the key factors that influence the performance of end-to-end deep learning pipelines for historical writer identification (HWI). Our contributions are as follows:
\begin{itemize}
    \item We explore multiple configurations of end-to-end deep learning pipelines to identify setups that offer a good balance between simplicity and performance.
    \item We evaluate the impact of various pre-processing strategies, including SIFT-based keypoint detection, document binarisation, and text region selection, on challenging historical document datasets.
    \item We investigate how different backbone architectures, loss functions, and post-processing techniques affect overall performance, and compare them with state-of-the-art, more complex approaches on widely used benchmark datasets.
\end{itemize}

Rather than aiming to find a single best-performing model, our goal is to understand which design choices contribute most to performance variation, particularly in zero-shot settings involving unseen writers and diverse document layouts. Our findings indicate that unstable feature representations, inconsistent patch-level outputs, and weak generalisation are key obstacles to building effective end-to-end HWI systems. These insights highlight the importance of integrating deep learning models with domain-specific pre- and post-processing steps to improve robustness in real-world scenarios.

\section{Related Works}
Several historical writer identification methods have been published over the years, proposing various image-processing techniques and using learning models. The process usually starts with pre-processing, in the form of binarization, noise removal, segmentation (letter, word, lines or paragraph level), and feature AOI selection (keypoints or specific contour in the image). The result of pre-processing then goes into feature extraction using image processing or a learning model, followed by post-processing to find similarity between documents.

\subsection{Image Processing Approach}
To capture writing style with non-standard patterns, various image processing techniques are used to transform the RGB/grayscale image into different vectors representing the meaningful abstraction on either a local or global scale, improving the aggregation of information that previously was hardly available from standard human observation. The local feature extraction focuses on capturing information containing the handwriting marker such as pen strokes, shape variations, and unique key points, using techniques such as HOG (Histogram of Oriented Gradients) \cite{Fiel2013, Peer2024}, oBIFs (Oriented Basic Image Features) \cite{Gattal2018, gattal2023new}, or SIFT (Scale-Invariant Feature Transform) \cite{xu2008handwriting,fiel2017icdar2017, lai2020encoding}. On the other hand, global feature extraction, like wavelet and GLCM (Grey-Level Co-Occurrence Matrix), captures global information in the form of large patterns and spatial relationships between lines on the entire image. With HWI tends to focus on the subtle writing details of each author, the local feature extraction approach is generally preferable, paired with image pre-processing and post-processing, like clustering, to reduce the amount of information due to other noise from the historical document.

\subsection{Deep Learning Approach}
The deep learning method generally uses different types of feature extraction backbone \cite{fagioli2023writer, bennour2024deep} to process information directly from the historical image. This approach gives more freedom compared to the image-processing approach since the model is not bound to the pre-determined features when finding the area of interest in the writing and learning the hidden writer characteristics from each author. Although this approach managed to get the best result in many classification tasks, using one for HWI is not easy due to the scarcity of labelled historical data needed to train the model. Even with the release of bigger HWI datasets \cite{fiel2017icdar2017, christlein2019icdar}, the combined number of data only covers slightly above 10000 writers, with the majority of them being part of the test set with minimal annotation and uneven class representation. Due to that, the robustness of image processing representation such as SIFT is still relevant, with some models using the technique as extra feature extraction during the training of a Convolutional Neural Network (CNN) \cite{chammas2020writer,christlein2022writer} or using transfer learning to expand the pre-trained network \cite{Cilia2019, Cilia2020}.


\section{Methods}

This study explores the fundamental design decisions that affect the performance of deep learning pipelines for historical writer identification (HWI). We frame the HWI task as a multi-stage process involving three core components: pre-processing, feature extraction (model selection and loss function), and post-processing. Each stage includes several possible configurations, and our objective is to systematically compare their effectiveness across multiple datasets under realistic conditions, including zero-shot evaluation with unseen writers. To structure this comparison, we define the main HWI pipeline as a transformation from the input document image \(I\) to a compact feature vector \(F \in \mathbb{R}^{N}\), where \(N\) depends on the chosen feature extraction and represents the dimensionality of the writer style representation. Pre-processing defines the process before the pipeline, while post-processing defines the process after the pipeline. In the ideal setup, the heavy lifting is done in the pipeline, while the rest is used as a way to improve the result. We tested each setup in different scenarios, highlighting aspects of each composition. Next, we discuss the details of each part in the HWI process.

\subsection{Pre-processing}
\begin{figure} 
\centering
\includegraphics[width=0.95\linewidth]{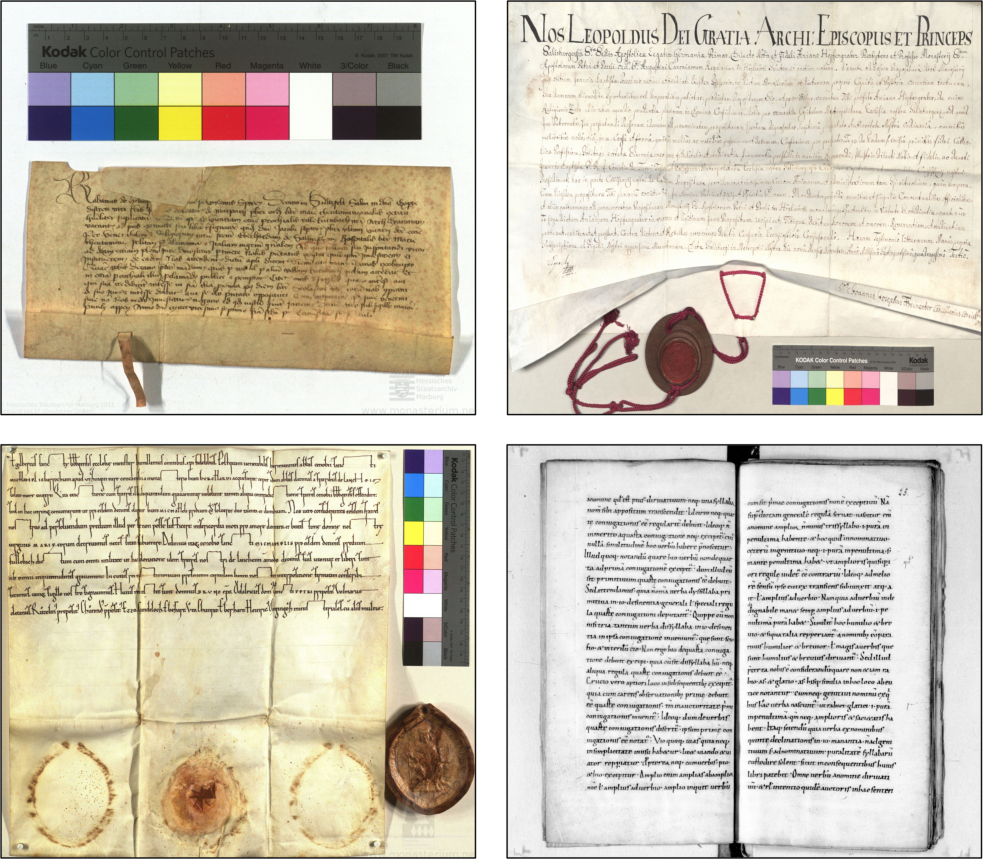}
\caption{The scan of a historical document image may contain noise that obstructs the writing information. While this might not be a problem for manual examination by an expert, it is a difficult challenge for automatic HWI, which needs to be addressed during pre-processing. The sample is taken from the ICDAR2019 HisIR  dataset\cite{christlein2019icdar}.}
  \label{fig:historical-document-quality} 
\end{figure}

We compare three different pre-processing setups and it's effect on the preparation of historical documents. While most document datasets are presented with necessary document preparation for the task, some still require hefty pre-processing due to the quality of the imaging process, as seen in Fig. \ref{fig:historical-document-quality}.

\subsubsection{Scale-Invariant Feature Transform (SIFT)}

SIFT is a widely used method for keypoint detection due to its invariance to scale, rotation, and minor affine transformations. It identifies local features around corners or textured regions that often correspond to unique writing styles. Keypoints are detected as local extrema in a scale-space using the Difference-of-Gaussians (DoG) function:

\begin{equation}
D(x, y, \sigma) = L(x, y, k\sigma) - L(x, y, \sigma),
\end{equation}

\begin{equation}
L(x, y, \sigma) = G(x, y, \sigma) * I(x, y)    
\end{equation}

\noindent where \(L\) is the image \(I\) convolved with a Gaussian kernel \(G\) at scale \( \sigma \), and \( k \) is a constant between successive scales. Candidates are selected by comparing each pixel with its 26 neighbours across adjacent scales. After applying contrast and edge filtering, descriptors are computed by extracting a \(16 \times 16\) patch around each keypoint, dividing it into \(4 \times 4\) subregions, and computing 8-bin orientation histograms, resulting in a normalised 128-dimensional vector.

As a handcrafted method, SIFT does not require training data and is well-suited for non-learning-based approaches such as Bag-of-Visual-Words (BoVW) or clustering. Adapting SIFT for deep learning method (with higher input resolution) relies on creating an image patch using keypoints as the centre, then using the patch area of the original image as the network's input instead of SIFT's \(128\)-dimensional feature vector.

\subsubsection{Historical Document Binarization}

Binarization is a crucial pre-processing step in historical document analysis, as it generates a text mask that helps HWI methods focus on textual regions while reducing background noise. Unlike classical thresholding techniques such as global methods (Otsu’s), local approaches (Sauvola’s), or the contrast and edge-response filtering used in SIFT, modern binarization methods are designed to learn text-specific patterns using training data. This makes them more effective in handling complex degradations, including faded ink, uneven lighting, or bleed-through from the reverse side of a manuscript (similar shape, wrong orientation).

Recent approaches look at binarization as a semantic segmentation task and apply deep learning models that can generalise better across diverse historical styles and conditions. In this study, we use an atrous binarization model \cite{rasyidi2021historical} trained on the DIBCO dataset series, which provides ground truth annotations for evaluating the quality of document binarization. Applying such a model to the original manuscript allows downstream processes, including SIFT-based or learning-based pipelines, to concentrate on more informative, text-dense regions of the document.

\subsubsection{Text-AOI Selection}

Finding a unique writing style in historical document writing is not easy, given the abundance of writing mixed in the document and the appearance of multiple noise patterns. Text-AOI (area of interest) selection utilises the combination of binarization and dilated connected component analysis to find the area in the document where the unique pattern might appear. The selection gets inspiration from the paleography process, in which the unique writing can be measured not only based on stroke curvature and letter forms, but also gaps and spaces between text groups. The Text-AOI selection that map \(I \rightarrow I_{\text{AOI}}\) can be represented as:

\begin{equation}
    I_{\text{AOI}} = I \cap (\text{BBOX}_{\text{top}}(\text{CCA}(D(B(I))))
\end{equation}

\noindent With the step-by-step process defined below:

\begin{enumerate}
    \item We process the original image \(I\) using a binarization model \(B\), to create a binary mask containing writing-like information. The model \(B\) should be trained using binarization datasets, enabling the identification of historical writing patterns from the rest of the document.
    \item The binary mask will undergo morphological dilation \(D\) with dilation factor \(d\) to close off the meaningful writing gaps (bigger gaps will be skipped since it is not counted as part of the writing pattern).
    \item The connected component analysis \(\text{CCA}\) is then applied to the dilated image to tie close writing groups together, then create a bounding box \(BBOX\) for each group.
    \item Create the bounding box ranking based on area and choose the top as the Text-AOI location.
    \item Crop original image \(I\) according to Text-AOI and use it for feature extraction training and testing.
\end{enumerate}

\subsection{Feature Extraction Model}

We evaluate the impact of different backbone architectures for feature extraction in historical writer identification (HWI) by comparing their performance during training and testing. The models considered in this study are:

\begin{itemize}
    \item Transformer models: SwinV2 \cite{liu2022swin} Small, Base, Large
    \item CNN models: ResNet18 \cite{he2016deep}, EfficientNetV2-S \cite{tan2021efficientnetv2}
\end{itemize}

The SwinV2 transformer approach is designed to find both local and global connections between patterns in the image, which connections define the whole premise of the writer's unique features. The CNN models look at the details of local connections, making them more similar to the HWI method using SIFT feature extraction, which focuses on the aggregation of pattern connections between multiple selected small areas.

For each backbone, an MLP head is attached to project the output feature maps into a fixed-dimensional feature vector \( F \in \mathbb{R}^{128} \) (for all Swinv2 and EfficientnetV2S), \( F \in \mathbb{R}^{256} \) (for Swinv2-Base training with cosine distance), and no MLP (for ResNet18 to mimic the setup of Siamese network approach \cite{Kumar2024SiameseWriterID}). The models are trained using Triplet loss to enable direct feature-level comparison between different architectures without relying on classification labels. The final similarity matrix is calculated by selecting patches of images, getting the feature representation for each patch, and using the average pooling of each feature dimension as the aggregated feature representation. One of the selected backbones is then further trained using ArcFace Loss after the initial Triplet loss training to let the model have a better feature separation.

\subsection{Loss Calculation}
\subsubsection{Triplet Loss}
To group the feature \(F \in \mathbb{R}^{N}\) of inputs \(I\) from the same writer, we employ the \textit{Triplet Loss} during the training of the feature extraction network. This encourages feature vectors from the same writer to be close in the latent space while pushing feature vectors from different writers apart. For that purpose, we sample patches from three different images \(I_a, I_p, I_n\), which are all forwarded into the proposed model to generate three feature vectors \(F_a, F_p, F_n)\):

\begin{itemize}
    \item \( F_a \): The feature vector of an \textit{anchor} input \( I_a \), which acts as the main image during data loading.
    \item \( F_p \): The embedding of a \textit{positive} input \( I_p \) (from the same writer as \( I_a \), chosen randomly from the dataset).
    \item \( F_n \): The embedding of a \textit{negative} input \( I_n \) (from a different writer than \( I_a \), chosen randomly from the dataset).
\end{itemize}

\noindent The goal is to ensure that the distance between \( F_a \) and \( F_p \) is smaller than the distance between \( F_a \) and \( F_n \) by at least a margin \( \alpha \).

The Triplet loss \(\mathcal{L_{\text{Triplet}}}\) is defined as:

\begin{equation}
    \mathcal{L}_{\text{Triplet}} = \max \big( 0, \| F_a - F_p \|^2 - \| F_a - F_n \|^2 + \alpha \big)    
\end{equation}

\noindent where:
\begin{itemize}
    \item \( \| F_a - F_p \|^2 \): Squared Euclidean distance between the anchor and positive vectors.
    \item \( \| F_a - F_n \|^2 \): Squared Euclidean distance between the anchor and negative vectors.
    \item \( \alpha > 0 \): A margin that separates positive and negative pairs, ensuring a minimum distance.
\end{itemize}

We additionally train one of the models using a variant of the Triplet loss based on cosine similarity, combined with an $L_2$ regularisation term. This setup allows us to assess whether using a different similarity metric affects the quality of the learned feature embeddings. The cosine-based Triplet loss is defined as:

\begin{equation}
\begin{split}
\mathcal{L}_{\text{Triplet}} =\ & \max\left(0,\ \cos(F_a, F_n) - \cos(F_a, F_p) + \alpha \right) \\
& + \lambda \left( \|F_a\|_2^2 + \|F_p\|_2^2 + \|F_n\|_2^2 \right)
\end{split}
\end{equation}

\noindent where \(cos\) is cosine similarity between two feature vectors, \( \lambda \) is Regularization weight to constrain the magnitude of the feature vectors, and \( \|F\|_2^2 \) is Squared L2 norm of a feature vector \( F \).

\subsubsection{ArcFace Loss}
To further improve the discriminative power of the feature vectors \( F \in \mathbb{R}^{N} \) during training, we also employ \textit{ArcFace loss} as an alternative to Triplet loss. ArcFace \cite{deng2019arcface} introduces an angular margin penalty in the classification layer, encouraging tighter intra-class feature distributions and larger inter-class margins by directly optimising for geodesic distance on a hypersphere. During training, each input feature \( F \) and its associated writer class label \( y \) are forwarded into an ArcFace classification head. The output logits are adjusted by adding an angular margin \( m \) to the target class before applying softmax.

Given a feature vector \( F \) and a class weight vector \( W_y \), the normalized cosine similarity is calculated as:

\[
\cos(\theta_y) = \frac{W_y^\top F}{\|W_y\| \|F\|}
\]

ArcFace modifies this similarity by introducing an additive angular margin:

\[
\cos(\theta_y + m)
\]

The ArcFace loss \(\mathcal{L}_{\text{arcface}}\) is then defined as:

\begin{equation}
    \mathcal{L}_{\text{arcface}} = -\frac{1}{N} \sum_{i=1}^N \log \frac{e^{s \cdot \cos(\theta_{y_i} + m)}}{e^{s \cdot \cos(\theta_{y_i} + m)} + \sum_{j \neq y_i} e^{s \cdot \cos(\theta_j)}}
\end{equation}

\noindent where:
\begin{itemize}
    \item \( N \): The number of samples in a mini-batch.
    \item \( \theta_y \): The angle between feature vector \( F \) and its corresponding class weight \( W_y \).
    \item \( m \): The additive angular margin to enforce a stricter classification boundary.
    \item \( s \): A scaling factor applied to the normalised cosine similarities to stabilise training.
\end{itemize}

\noindent By optimising this loss, the model learns features that are not only separable but also better aligned for retrieval and clustering tasks common in HWI, resulting in a feature extraction that produces a well-separated representation for each writing style.

\subsection{Post-Processing}

Post-processing is an optional step to consolidate these features into a document-level embedding. We evaluate two primary strategies: pooling-based aggregation and PCA dimension reduction.

\subsubsection{Pooling Aggregation}
Mean pooling is applied to combine the features of multiple patches into a single feature vector representing the entire document. Given a set of patch embeddings \( \{F_1, F_2, \ldots, F_k\} \), we perform pooling across the feature dimensions, followed by the distance calculation using Euclidean distance or cosine similarity, based on the network setup.

\subsubsection{PCA Dimensional Reduction}
We also apply PCA on the feature vector before the pooling aggregation to see the effect of different dimensional reduction on the similarity result. Since not all the embedding dimension has a similar degree of representation, PCA is known to have a positive impact on reducing redundancy and noise from the original vector.




\section{Experiments}

\subsection{Network Architecture} 
We are using two types of network architecture in this research: an encoder-decoder binarization model and a writing feature extraction model. 

\subsubsection{Historical Document Binarization Model}
The binarization model uses the atrous binarization model proposed in \cite{rasyidi2021historical} using the atrous ResNet18 backbone. We chose this architecture due to the use of dilated convolution and atrous spatial pyramid pooling (ASPP), which adds more flexibility to work with complex writing styles and the ability to retain the details of stroke and letter shape from the original document. We train the model using DIBCO'09 to DIBCO'14 datasets (including H-DIBCO for handwriting data) with pseudo-F loss \(\mathcal{L}_{fps}\) following the setup from the original paper, then validate the result with the DIBCO'16 dataset.

\subsubsection{Writing Feature Extraction Model}
The transformer architecture uses pre-trained SwinV2 models combined with a three-layer MLP head to learn various writing feature representations in the area. We use Parameter Efficient Fine Tuning (PEFT) using LoRA \cite{hu2021lora} to work with the selective weight adjustment to 'query', 'key', and 'value' parts of the SwinV2's self-attention mechanism. With this, all the trainable parameters of the pre-trained SwinV2 will be frozen, and the LoRA will attach trainable low-rank matrices, enabling fine-tuning with only a fraction of the computational cost compared to doing it with a full model. As for the pre-trained CNN architecture (EfficientNetV2S and ResNet18), we freeze several early layers of the network to focus on the fine-tuning of the later part of the pipeline. ResNet18 take SIFT-centred image crop as input, while the other model uses random patches. All of the networks are trained using Triplet Loss, with a selected few further fine-tuned using ArcFace.

\subsection{Dataset Preparation}
We use three datasets with different characteristics in our experiment to test the proposed model in a variety of scenarios.

\subsubsection{ICDAR2013-WI} Is a writer identification dataset \cite{louloudis2013icdar} containing 1400 image from 250 distinct writer. The dataset is divided into experimental (400 images from 100 writers) and benchmarking (1000 images from 250 writers). Since the dataset contains well-processed writing lines on a white background, we use the data to measure the baseline performance of our feature extraction network, which is trained using the experimental set and validated/tested using the benchmarking set. No pre-processing or Text-AOI Selection is needed, except for image resizing with factors \(r = 0.5\) to standardise the image with our feature extraction input. This dataset is used for the comparison of different feature extraction backbone for solving WI task and the comparison of our best approach with other benchmarked methods.

\subsubsection{ICDAR2017-HistoricalWI} Is a historical writer identification dataset \cite{fiel2017icdar2017} containing document scans that are presented in a training set (1182 images from 394 writers) and a test set (3600 images from 720 writers). Each writer only appears on one set, making the dataset have a total of 4782 images from 1094 district writers. In terms of quality, both training and test sets present clear pages of historical document scans without meaningful distortion or bad scan quality. Text-AOI is selected from the dataset, using the dilation factor \(d = (30, 30)\) pixels to create a sub-document with the most textual information. The dataset is used for training and testing for method comparison.

\subsubsection{ICDAR2019-HisIR} Is a historical dataset that focuses on document retrieval based on writing style \cite{christlein2019icdar}. This dataset presents a challenge compared to other HWI datasets since this contains raw document scans on the test sets aside from the corrected clean pages scan in the previous dataset. Due to having a high resolution, we process the data by using the resizing factors \(r = 0.5\) and \(r = 0.25\) depending on the size of the image scans before processing them for Text-AOI selection. Due to the complex representation of the scanned documents, we find that using dilation factor \(d = (25, 25)\) balances the process of working with a small-sized text area while keeping the unrelated text/noise away. Manual adjustment is done for selected miss-segmented images. This dataset presents the case where we can see the importance of using Text-AOI Selection for HWI, and used for comparing the difference in pre-processing methods.

\subsection{Hyperparameters}
The binarization model is trained using an SGD optimizer with the learning rate \(lr = 0.001\) and \(momentum = 0.9\) following the original implementation. The SwinV2 model use LoRA with parameters \(r = 32\), \(\alpha = 64\), and \(dropout = 0.05\). All feature extraction training was conducted using an AdamW optimizer with a learning rate of \(0.0001\) using \(\mathcal{L}_{\text{Triplet}}\) with \(margin=1.0\) and \(p=2\). The SwinV2-Base model is fine-tuned using the ArcFace loss head until it reaches loss convergence.



\section{Results and Discussions}

\subsection{Pre-processing Result Comparison}
\begin{figure*} 
\centering
\includegraphics[width=0.95\linewidth]{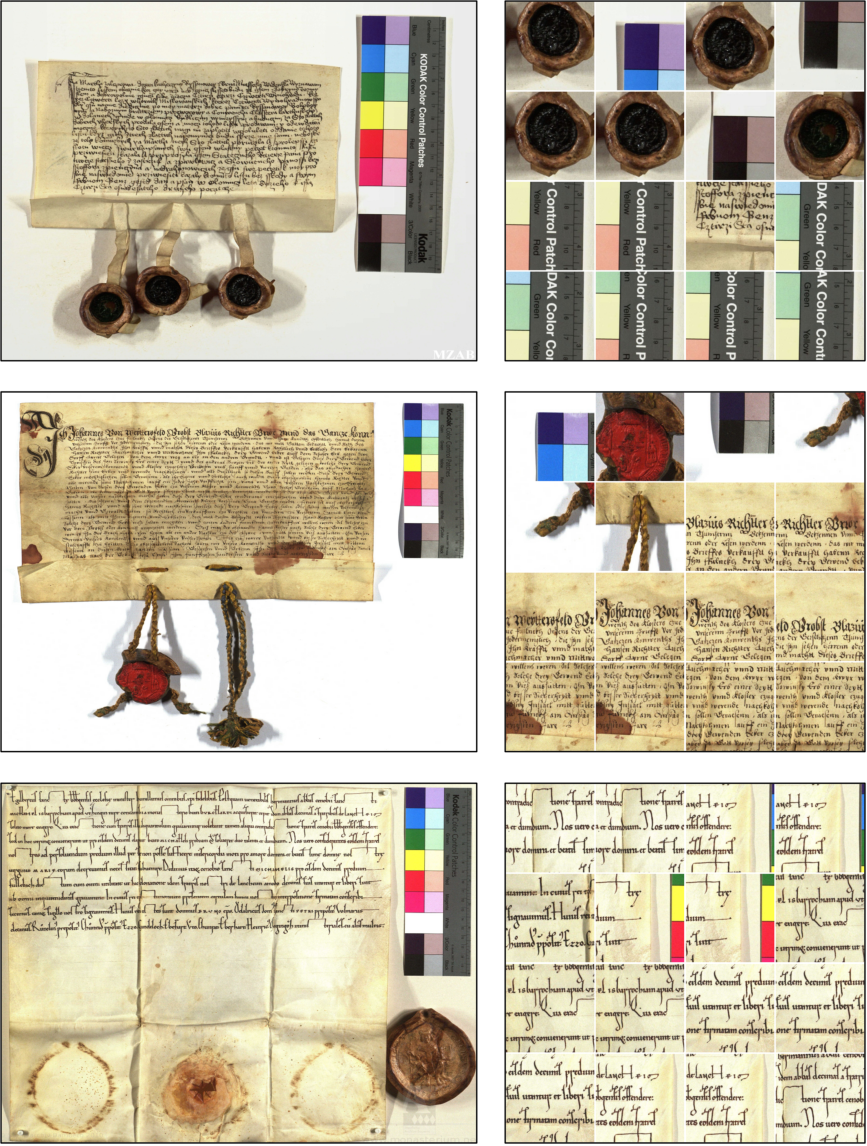}
\caption{Sample patches taken from the historical document scan (left) using SIFT keypoint as anchor (right). With SIFT not understanding the general writing pattern, it may result in incorrect keypoint selections when used in an image with heavy-textured noise. The sample is taken from the ICDAR2019 HisIR  dataset\cite{christlein2019icdar}.}
  \label{fig:sift-patches} 
\end{figure*}

We test different pre-processing methods on the HisIR test data to evaluate how well each approach handles difficult historical documents. Sample patches centred on SIFT anchors are shown in Fig.~\ref{fig:sift-patches}. Based on the examples, we observe that SIFT struggles to accurately capture text regions in images with complex noise. Since SIFT lacks any understanding of general text patterns, noisy artefacts are often selected as top candidates instead of actual text. Moreover, because deep learning models typically use larger patches than the area around individual SIFT keypoints, closely spaced keypoints can lead to highly similar input patches, reducing the diversity needed for effective training. In images with thousands of detected keypoints, randomly sampling from them is prone to producing results similar to naive random patch selection.

\begin{figure*} 
\centering
\includegraphics[width=0.95\linewidth]{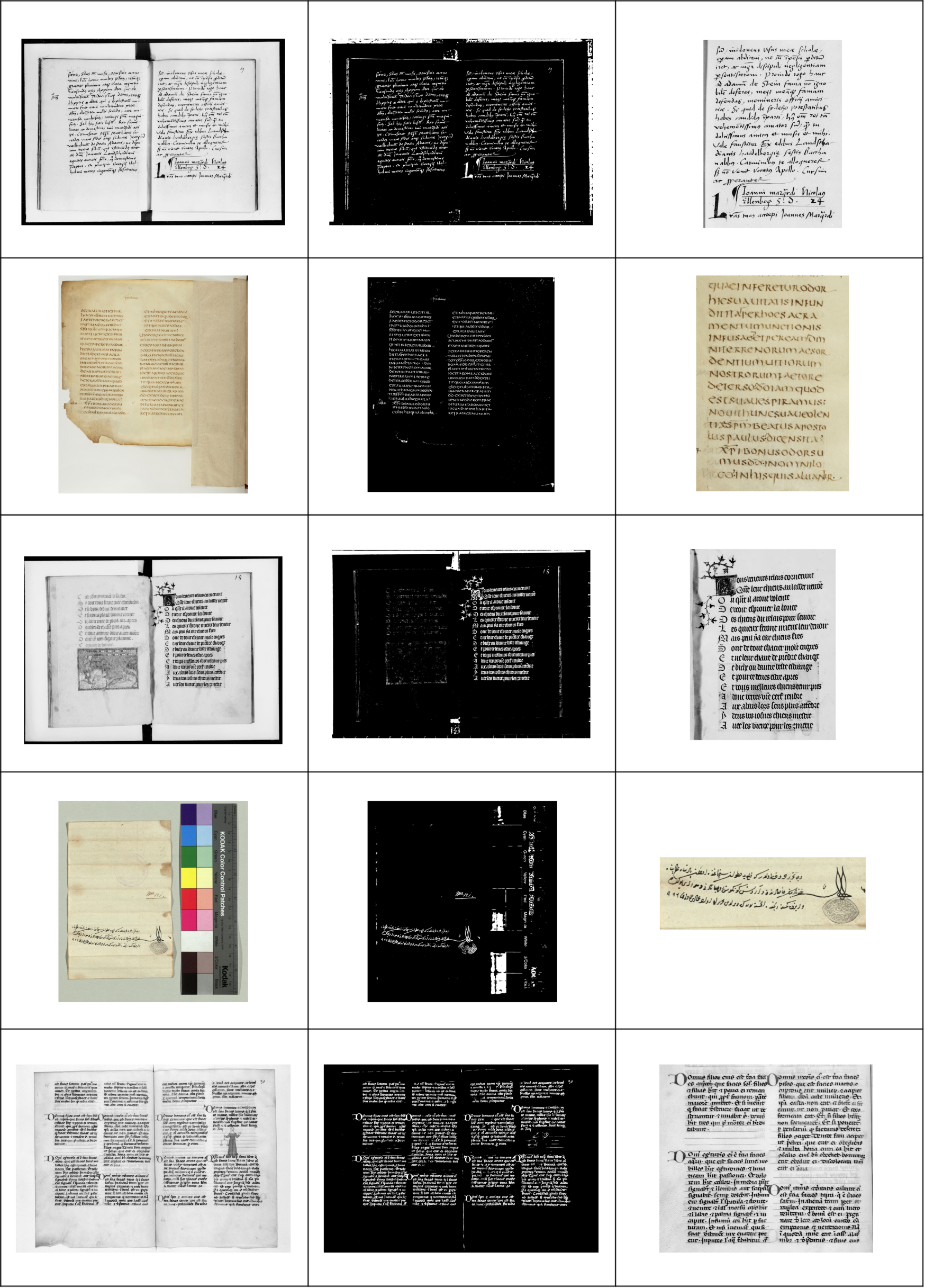}
\caption{The process of transforming a historical document scan with Text-AOI selection, showing the transformation from the original image (left), into binarization output (middle), ending with the Text-AOI image (right). The data is sampled from ICDAR2019 HisIR dataset \cite{christlein2019icdar}.}
  \label{fig:binarization-and-aoi-selection} 
\end{figure*}

A comparison between document binarisation and Text-AOI selection is shown in Fig.~\ref{fig:binarization-and-aoi-selection}. While binarisation can help isolate text, the results often still contain artefacts, particularly when the quality of the scanned document differs from the binarization training data. The Text-AOI selection step introduces an additional process to filter out less relevant regions, focusing the result on areas of the document that are more likely to contain distinctive writing patterns. When combined with random patch selection, this approach helps the network concentrate on meaningful visual patterns during training, as opposed to sampling indiscriminately from the entire document.

\subsection{Comparing Different Architectures and Post-processing}
\begin{figure*}
    \centering
  \subfloat{%
       \includegraphics[width=0.9\linewidth]{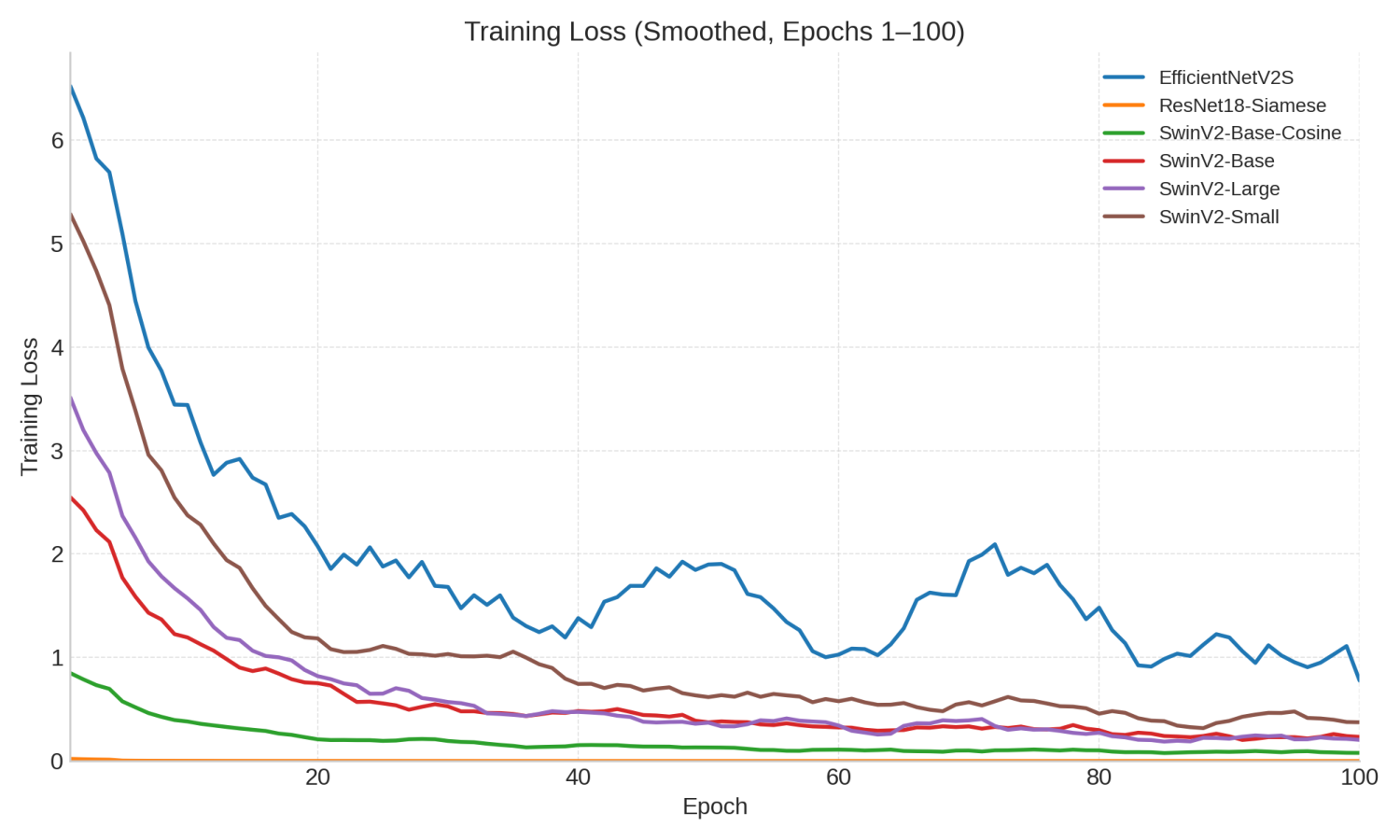}}
    \hfill \\
  \subfloat{%
        \includegraphics[width=0.9\linewidth]{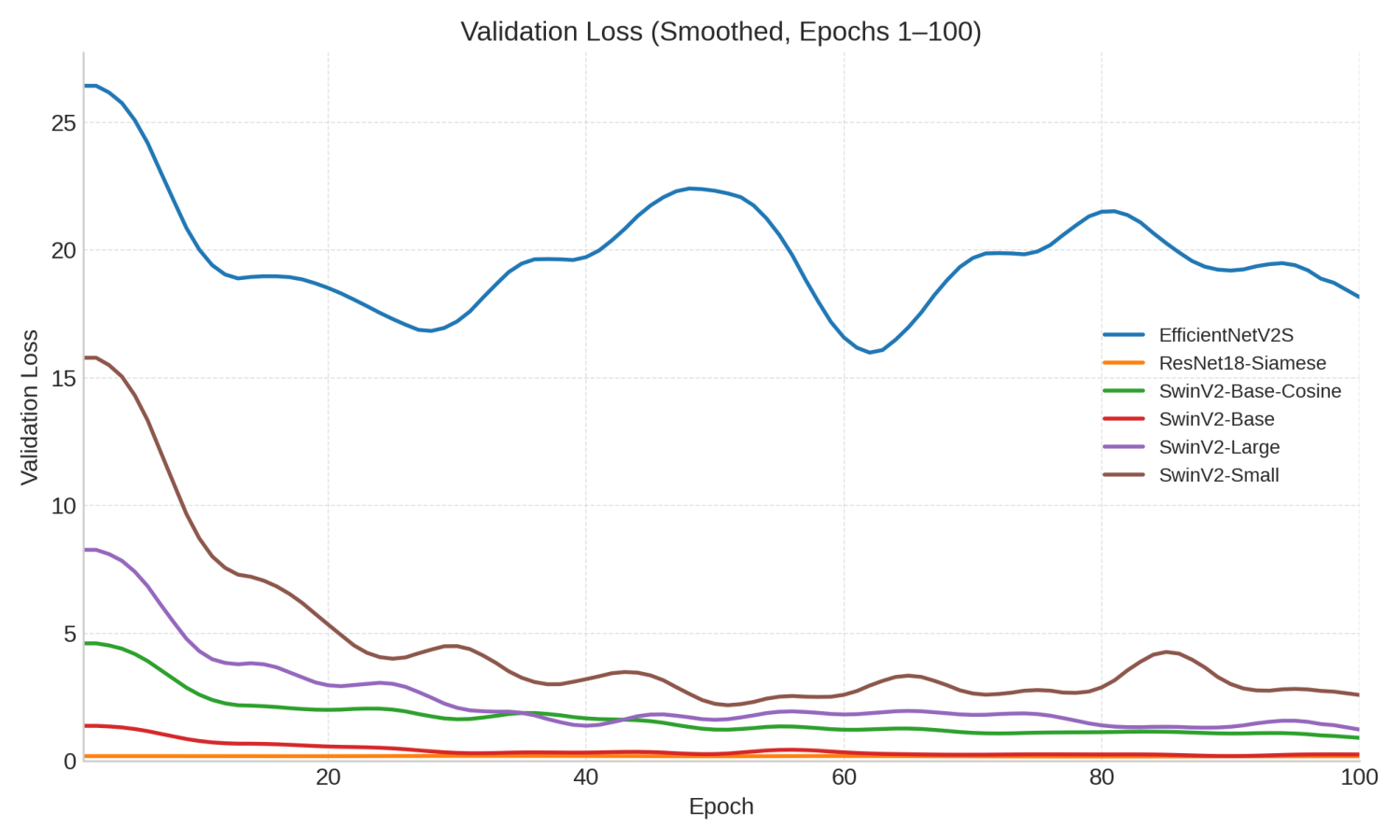}} 
        \\
  \caption{Training and validation loss comparison between different backbone setups on ICDAR2013-WI dataset. We can see that the SwinV2-Base model has better training and validation trends compared to the other models, with the version with L2 regularization and cosine distance having a more constant loss reduction.}
  \label{fig:backbone-loss-comparison} 
\end{figure*}

The comparison of training and evaluation loss of different backbone architectures is shown in Fig.~\ref{fig:backbone-loss-comparison}, each using a similar training setup and dataset. We can see that while ResNet18 has a constantly lower loss in both training and validation, the validation is stagnant and has a similar trend to the one that happened to EfficientNetV2, meaning that they are not properly learning the general pattern separation during training. On the other hand, SwinV2-Base achieve a steadier trend in loss decline in both training and validation, with the version with cosine Triplet loss performing the best due to the use of L2 regularization. Using the model on the test data with mean pooling as post-processing produces the accuracy displayed in Table~\ref {tab:icdar2013-wi-backbone-comparison}. Swinv2-Large managed to top the raw accuracy compared to the other models, but it is costly to train and takes more resources to be fine-tuned further. Applying PCA dimensional reduction for post-processing generally does not help in this case, and only provides a slight improvement on the model with L2 normalisation, as seen in Table~\ref{tab:icdar2013-wi-comparison_pca}. Based on the result, we can also see that the SwinV2 transformer model achieves relatively better results compared to its CNN counterpart, since it takes both global and local patterns from document patches, in contrast to only focusing on local stroke patterns that do not always appear in the input image patches.

\subsection{The Effect of ArcFace on Feature Extraction}
Fine-tuning the SwinV2-Base model from the previous training using ArcFace produces strong performance that is comparable to other, more complex top-performing methods, while maintaining a simple end-to-end deep learning pipeline. This result is consistently observed in both the standard WI dataset, as shown in Table~\ref{tab:icdar2013-wi-comparison_top1_top5}, and the HWI dataset, as shown in Table~\ref{tab:icdar2017-wi-comparison_top1_top5_mAP}. Although ArcFace is primarily designed for classification tasks, it outperforms Triplet Loss in extracting discriminative features from the backbone. This is likely because Triplet Loss optimises only a limited number of triplets at a time, which makes it difficult to tune effectively in zero-shot settings with high variability and inconsistent handwriting patterns, as often seen in HWI. In contrast, ArcFace enforces angular margin constraints across all classes during training, encouraging better feature separation and leading to more generalisable representations, even for unseen writers. This comes with the price of slower training, which makes this approach more suitable for fine-tuning scenarios.

\begin{table*}[ht]
\centering
\captionsetup{justification=centering, width=0.75\linewidth}
\caption{Comparison of Top-1 and Top-5 retrieval accuracy between different backbones trained using Triplet Loss.}
\begin{tabular}{|c|c|c|}
\hline
\textbf{Backbone} & \textbf{Top-1 Accuracy (\%)} $\uparrow$ & \textbf{Top-5 Accuracy (\%)} $\uparrow$ \\
\hline
Swinv2-Small & 31.50 & 46.90 \\
Swinv2-Base & 33.60 & 53.20 \\
Swinv2-Base + Cosine & 39.80 & 58.10 \\
Swinv2-Large & 40.20 & 61.10 \\
EfficientNetV2S & 7.10 & 21.20 \\
ResNet18 \cite{Kumar2024SiameseWriterID} & 17.40 & 33.50 \\
\hline
\end{tabular}
\label{tab:icdar2013-wi-backbone-comparison}
\end{table*}

\begin{table*}[ht]
\centering
\captionsetup{justification=centering, width=0.75\linewidth}
\caption{Comparison of Top-1 result between different PCA dimension for ICDAR2013-WI Dataset. Higher values indicate better performance.}
\begin{tabular}{|c|c|c|c|c|}
\hline
\textbf{Model} & \textbf{Original} & \textbf{PCA-128 (\%)} $\uparrow$ & \textbf{PCA-64 (\%)} $\uparrow$ & \textbf{PCA-32 (\%)} $\uparrow$ \\
\hline
SwinV2-Small        & 31.50 & 28.50 & 28.50 & 28.50 \\
SwinV2-Base         & 33.60 & 32.70 & 32.70 & 32.00 \\
SwinV2-Base Cosine  & 39.80 & 40.40 & 40.20 & 39.20 \\
SwinV2-Large        & 40.20 & 38.20 & 38.10 & 37.60 \\
\hline
\end{tabular}
\label{tab:icdar2013-wi-comparison_pca}
\end{table*}

\begin{table*}[ht]
\centering
\captionsetup{justification=centering, width=0.75\linewidth}
\caption{Comparison of Top-1 and Top-5 retrieval accuracy between different leading methods for ICDAR2013-WI Dataset. Higher values indicate better performance.}
\begin{tabular}{|c|c|c|c|}
\hline
\textbf{Name} & \textbf{Method} & \textbf{Top-1 Accuracy (\%)} $\uparrow$ & \textbf{Top-5 Accuracy (\%)} $\uparrow$ \\
\hline
Rank-1 \cite{louloudis2013icdar} & Contour Gradient & 95.60 & 98.60 \\
\cite{rasoulzadeh2022writer} & NetVLAD + Re-Ranking & 98.70 & - \\
\cite{koepf2022writer} & ViT + KNN &  97 & 98.6 \\
Ours & SwinV2 + ArcFace & 97.40 & 99.25 \\
\hline
\end{tabular}
\label{tab:icdar2013-wi-comparison_top1_top5}
\end{table*}

\begin{table*}[ht]
\centering
\captionsetup{justification=centering, width=0.75\linewidth}
\caption{Comparison retrieval accuracy between different leading methods for ICDAR2017-HistoricalWI Dataset. Higher values indicate better performance.}
\begin{tabular}{|c|c|c|c|c|}
\hline
\textbf{Name} & \textbf{Method} & \textbf{Top-1 (\%)} $\uparrow$ & \textbf{P@2 (\%)} $\uparrow$ & \textbf{mAP (\%)} $\uparrow$ \\
\hline
\cite{gattal2023new} & oBIFs columns histogram & 76.4 & 68.4 & 55.6 \\
\cite{Jordan2020} & Pair \& Triple-SVM & 89.43 & - & 78.20\\
\cite{christlein2017unsupervised} & Clutering + CNN + Fisher + SVM &  88.9 & 84.8 & 76.2 \\
Ours & SwinV2 + ArcFace & 97.15 & 96.08 & 42.16 \\
\hline
\end{tabular}
\label{tab:icdar2017-wi-comparison_top1_top5_mAP}
\end{table*}

\section{Conclusion and Future Works}
In this paper, we presented a study and analysis of key factors that influence the performance of end-to-end deep learning approaches for historical writer identification (HWI). Our investigation into pre-processing methods showed that combining document binarisation with Text-AOI selection effectively isolates relevant text regions from background noise, allowing random patch selection to be safely applied during training of the feature extraction network. Among the models evaluated, the SwinV2-Base architecture with L2 regularisation achieved the highest accuracy. When paired with ArcFace loss, this setup matched the performance of more complex state-of-the-art methods on both the WI and HWI datasets, while maintaining a simpler end-to-end structure. However, this improvement comes at the cost of longer and more resource-intensive training compared to models trained with Triplet Loss. As future work, exploring more efficient fine-tuning strategies may help reduce this cost, especially as larger historical writer identification datasets become available.

\bibliography{reference.bib}
\bibliographystyle{IEEEtran}

\end{document}